# Explainable Artificial Intelligence (XAI): Precepts, Methods, and Opportunities for Research in Construction

Peter E.D. Love[a], Weili Fang[b,c, *], Jane Matthews[d] , Stuart Porter[e], Hanbin Luo[f], Lieyun Ding[g]

[a] School of Civil and Mechanical Engineering, Curtin University, GPO Box U1987, Perth, Western Australia 6845, Australia, Email: p.love@curtin.edu.au

[b] Department of Civil and Building Systems, Technische Universität Berlin, Gustav-Meyer-Allee 25, 13156 Berlin, Germany, Email: weili.fang@campus.tu-berlin.de

[c] School of Civil and Mechanical Engineering, Curtin University, GPO Box U1987, Perth, Western Australia 6845, Australia, Email: weili.fang@curtin.edu.au

[d] School of Architecture and Built Environment, Deakin University Geelong Waterfront Campus, Geelong, VIC 3220, Australia, Email: jane.matthews@deakin.edu.au

[e] School of Civil and Mechanical Engineering, Curtin University, GPO Box U1987, Perth, Western Australia 6845, Australia, Email: stuart.porter@curtin.edu.au

School of Civil Engineering and Mechanics, Huazhong University of Science and Technology, Wuhan, 430074, China, Email: luohbcem@hust.edu.cn

School of Civil Engineering and Mechanics, Huazhong University of Science and Technology, Wuhan, 430074, China, Email: dly@hust.edu.cn

[*] Corresponding Author: Weili Fang, Email: weili.fang@campus.tu-berlin.de

# Explainable Artificial Intelligence (XAI): Precepts, Methods, and Opportunities for Research in Construction

**Abstract**

Deep learning (DL) and machine learning (ML) are major branches of Artificial Intelligence. The 'black box' nature of DL and ML makes their inner workings difficult to understand and interpret. The deployment of explainable artificial intelligence (XAI) can help explain *why* and *how* the output of DL and ML models are generated. As a result, an understanding of the functioning, behavior, and outputs of models can be garnered, reducing bias and error and improving confidence in decision-making. Despite providing an improved understanding of model outputs, XAI has received limited attention in construction. This paper presents a narrative review of XAI and a taxonomy of precepts and methods to raise awareness about its potential opportunities for use in construction. It is envisaged that the opportunities suggested can stimulate new lines of inquiry to help alleviate the prevailing skepticism and hesitancy toward AI adoption and integration in construction.

## 1.0 Introduction

Artificial Intelligence (AI) is the mimicking of human intelligence processes by machines, especially computer systems. Two major offshoots of AI are Deep Learning (DL) and Machine Learning (ML), which can provide businesses with a wide range of benefits if applied appropriately (Hussain *et al*., 2021).

The outputs generated by ML and DL models are produced by black boxes that are impossible to interpret (Gunning *et al*., 2018). In computer science, the term black box refers to a data-driven algorithm that produces useful information without revealing any information about its internal workings (Sokol and Flach, 2022). Consequently, AI models must be continuously monitored and managed to explain their use and the outputs of algorithms (Arrieta *et al*., 2021).

The difference between DL and ML needs to be highlighted at this juncture. In short, ML is a subfield of AI and can learn and adapt without following explicit instructions by using algorithms and statistical models to analyze and draw inferences from patterns of data. In contrast, DL is a subfield of ML whereby artificial neural networks (ANN) form the basis of its developed algorithms. The ANN comprises multiple layers that process data and extract progressively higher-level features.

Data-driven DL and ML models are good at unearthing associations and patterns in data, but causality cannot be guaranteed. Prior knowledge needs to be considered to infer causal relationships. The discovered associations emerging from AI-based models may be completely unexpected, not interpretable nor explainable (Longo *et al*., 2020). Thus, with the rapid advancements in AI, people have begun to question the risks associated with its use and have sought to comprehend how an algorithm arrives at a solution (Flok *et al*., 2021).

As Anand *et al*. (2020) cogently point out, DL and ML are akin to black magic as their underlying structures are complex, non-linear, and difficult to interpret and explain to lay people. Even those who create the DL and ML algorithms often cannot explain their internal mechanisms and how a solution is derived. The creators do not know how DL and ML models learn and make predictions (Arrieta *et al*., 2020; Amarasinghe *et al*., 2021; Langer *et al*., 2021; McNamara, 2022). Accordingly, entrusting important decisions to a system that cannot explain itself is naturally risky for decision-makers (i.e., end-users) (Adadi and Berrada, 2018). With this in mind, Explainable AI (XAI) has emerged to help users understand how an AI-enabled system results in specific outputs.

Simply put, the purpose of an XAI system is to make its behavior more intelligible to humans by providing explanations so that end-users can comprehend and trust the outputs generated by ML and DL algorithms (Gunning *et al*., 2018). That said, Arrieta *et al*. (2020) provide the following definition for XAI: "Given an audience, an [XAI] is one that produces details or reasons to make its functioning clear or easy to understand" (p.85). While this definition is used in this paper, the fact remains that there is no consensus on what constitutes XAI (Murray, 2021).

Regardless of an agreed definition, XAI systems should be able to explain their capabilities, "what it has done, what it is doing, and what will happen" and then reveal the relevant information they are acting on (Gunning *et al*., 2018: p.1). To this end, XAI incorporates issues associated with *interpretability and transparency* (Gilpin *et al.,* 2018; Arrieta *et al*., 2020; Flock *et al*., 2021; Hussain *et al*., 2021; Tjoa and Guan, 2021).

Despite the increasing shift toward the use of XAI in areas such as law (Clarke, 2019) and medicine (Lamy *et al.*, 2019), reviews of AI in construction surprisingly have not given attention to the subject (Darko *et al*., 2020; Abdirad and Mathur, 2021; Xu *et al*., 2021) (Debrah *et al*., 2022; Pan and Zhang, 2022; Zhang *et al*., 2022), though Abioye *et al*. (2021) do refer to it in passing. Though, there appears to be some traction in addressing XAI in construction, as evident in the works of Gao *et al*. (2023) and Wang *et al*. (2023). Apart from these latest publications, the literature in construction still remains relatively silent on the topic of XAI.

Indeed, construction organizations have been hesitant to embrace AI, not for the reason that they are unwilling or averse to its adoption but because its autonomous decisions and actions are unexplainable (Matthews *et al*., 2022). Confidence and trust in the decisions become doubtful as a prediction is unexplainable. The potential of XAI is profound for construction (e.g., reduced impact of model bias, the building of trust, development of actionable insights, and mitigating the risk of unintended outcomes) as it can be used to explain decisions and competencies and also have a social role. In this instance, social roles not only include learning and explaining to individuals but also coordinating with other agents to connect knowledge, develop cross-disciplinary insights and shared interests, and draw on previously discovered knowledge to accelerate further discovery and application (Gunning *et al*., 2018)

The motivation for this paper is *not* to provide a review of AI in construction. The literature is replete with such works (e.g., Darko *et al*., 2020; Debrah *et al*., 2022; Zhang *et al*., 2022). Instead, the paper aims to fill the void that has been overlooked in such studies by examining precepts and methods of XAI and identifying opportunities for research. It should be acknowledged that reviews of XAI outside of the domain of construction are plentiful as it is an area of growing importance (Arrieta *et al*., 2020; Longo *et al*., 2020; Flock *et al*., 2021; Sokol and Flach, 2022). However, this paper aims to stimulate interest in XAI and encourage a shift away from traditional AI research in construction so algorithmic black boxes can be questioned and understood by end-users.

This review paper is not systematic due to the novelty of XAI in construction. Only a limited number of studies have utilized XAI (Miller, 2019; Naser, 2021; Geetha and Sim, 2022; Gao *et al*., 2023; Wang *et al*., 2023). So, instead, a narrative review is adopted and, in doing so, provides a comprehensive examination, and objective analysis of current knowledge about XAI positioned within the context of construction.

The paper commences by introducing the precepts of XAI and proposes a conceptual taxonomy to navigate the emergent and complex literature (Section 2). Then, the methods used to support XAI (i.e., transparent and opaque models and post-hoc explainability) are examined (Section 3). As XAI is an emerging topic for construction, opportunities for future research are proposed (Section 4) before concluding the paper (Section 5).

## 2.0    Explainable AI

An overview of how AI is applied in construction compared to XAI is presented in Figure 1 to provide a context for the paper and subsequent review. Underpinning XAI is the precepts of explainability and interpretability (Gilpin *et al*., 2018; Arrieta *et al*., 2020; Longo *et al*., 2020). Both these precepts are often used interchangeably, but they have distinct differences (Arrieta *et al*., 2020). In a nutshell, explainability refers to understanding and clarifying the internal mechanics of a DL or ML model so they can be explained in human terms. Likewise, interpretability focuses on predicting what will happen, given changes to the inputs of a model and algorithmic parameters. Thus, interpretability aims to observe the cause-and-effect workings of a model.

Of note, trustworthiness has also been associated with explainability and interpretability. However, how trust is incorporated into the vast array of existing ML models is ambiguous (Gilpin *et al.,* 2018). Reinforcing this ambiguity, Lipton (2018) questions the meaning of trust, likening it to simply having confidence that a model is performing well. So, if a model is demonstrated to be accurate, it should be trustworthy. Thus, in this case, interpretability serves no purpose. Furthermore, trust can be subjective, which suggests that a person may accept a well-understood model even though it serves no purpose. This issue is associated with completeness, which will also be addressed below.

Meanwhile, a taxonomy of XAI derived from the literature is presented in Figure 2 to help understand the relations between the precepts and methods and the structure of the review (Arrieta *et al.*, 2020; Angelov *et al.*, 2021; Belle and Papantonis, 2021; McDermid *et al.*, 2021; Langer *et al.*, 2021; Vilone and Longo, 2021; Minh *et al.*, 2022; Speith, 2022; Yang *et al.*, 2022). The following sections of this paper examine the precepts of explainability and interpretability.

## 2.1    *Explainability*

The historical notion of scientific explanation has been the subject of intense debate in the arenas of science and philosophy (Woodward, 2003). Aristotle suggested that knowledge "becomes scientific when it tries to explain the causes of *why*" (Longo *et al.*, 2020: p.3). Markedly, questions about what makes a good enough explanation have also been a constant source of debate (Thagard, 1978; Caliskan *et al.*, 2017). But fundamentally, what makes an explanation understandable depends on the questions being asked (Bromberger, 1992).

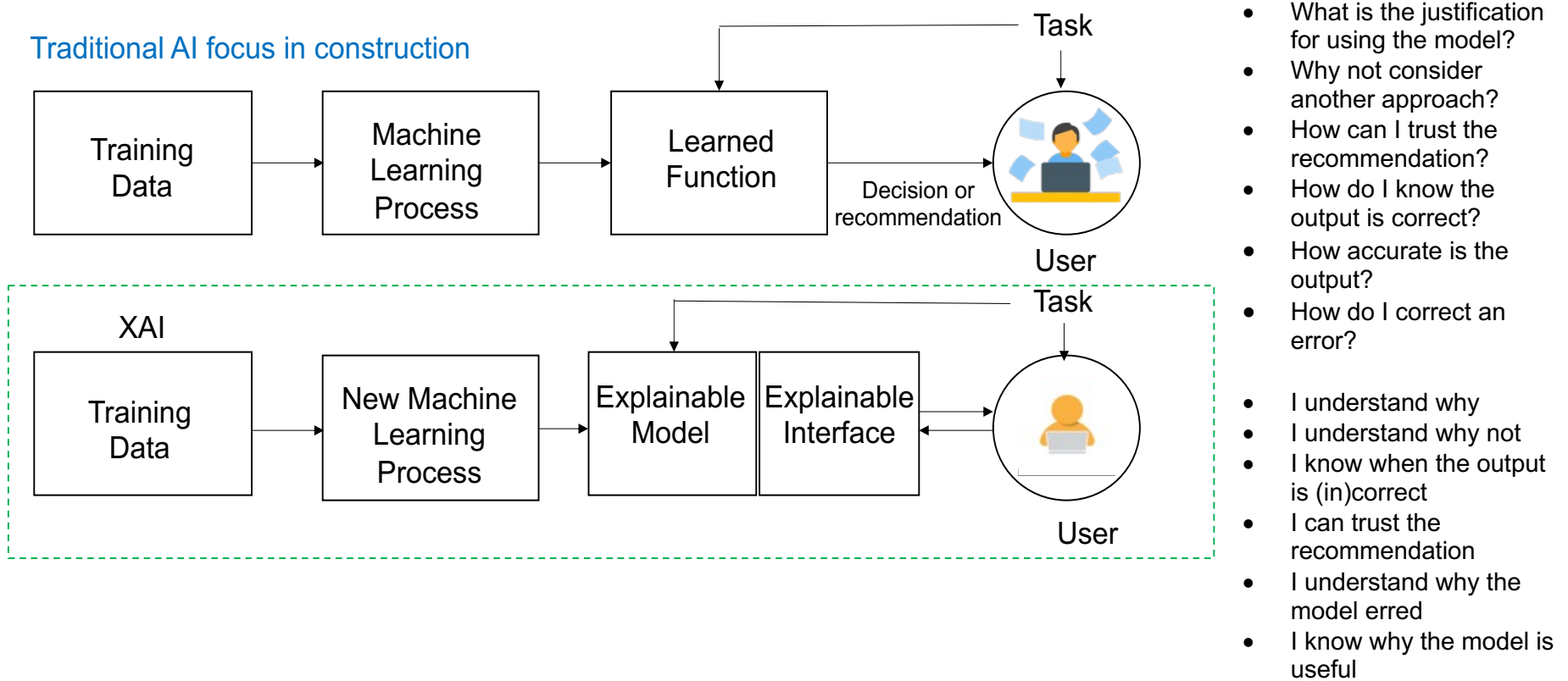

Adapted from McNamara (2022)

Figure 1. A comparison between traditional the AI focus in construction and XAI

In the case of XAI, explainability describes the active properties of a learning model based on *posteriori* knowledge, which aims to clarify its functioning (Flock *et al*., 2021). Here, explainability is subjective as end-users are placed in a position to determine whether the explanation provided is believable (Arrieta *et al*., 2020). To ensure the authenticity of outputs generated from models, there is a need to explain (McNamara, 2022):

- What data went into a training model and why? How was fairness assessed? What effort was made to reduce bias? (*Explainable data*)

- What model features were activated and used to achieve a given output? (*Explainable predictions*)

- What individual layers make up the model, and how do they result in an output (*Explainable algorithms*)

Notably, the outputs of studies using DL and ML in construction are seldom questioned and validated by end-users in a real-life context, particularly those related to computer vision (Love *et al*., 2022a). Moreover, the usefulness of a DL and ML solution for a problem that addresses an industry need within the academic construction literature is hardly ever examined.

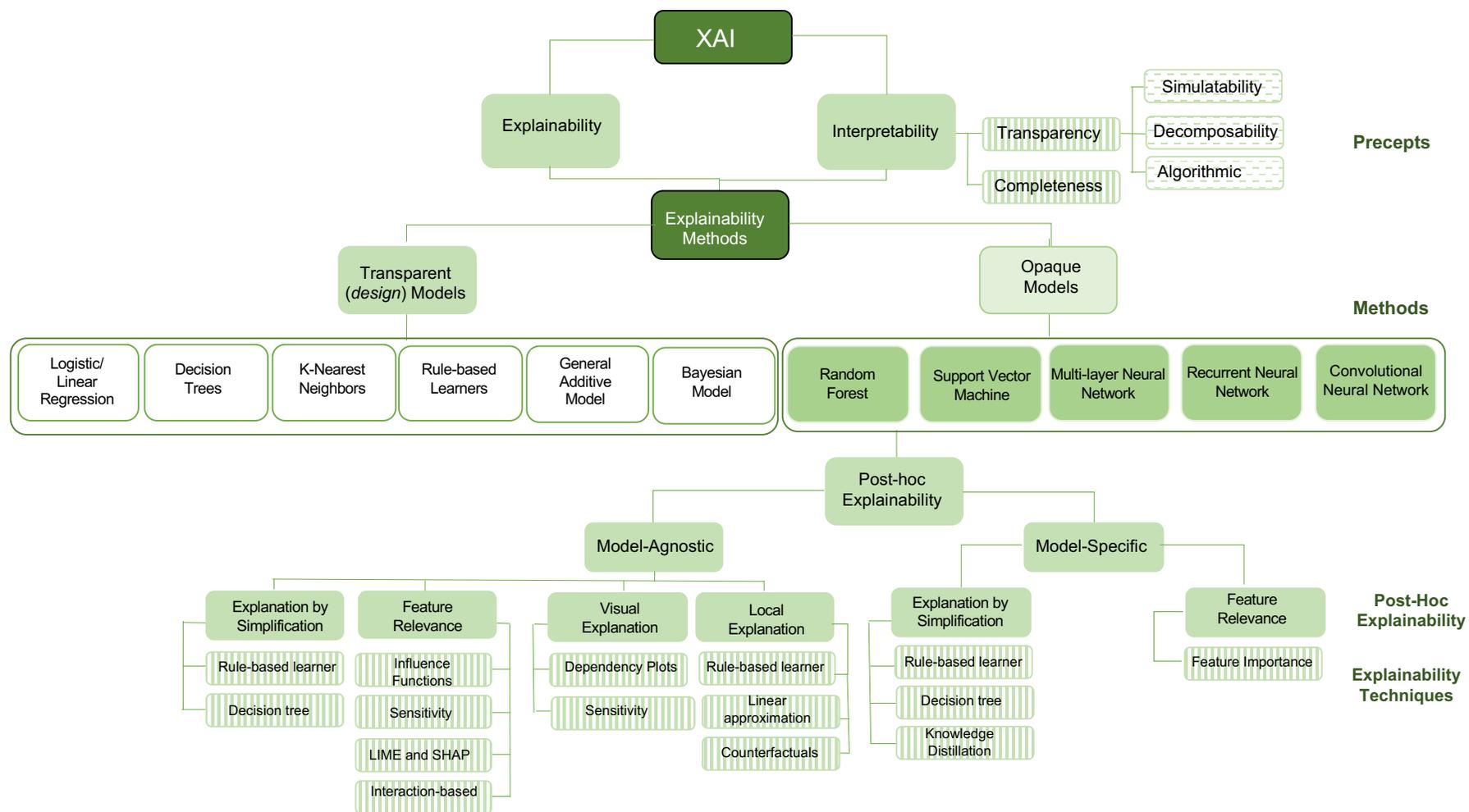

Adapted from Angelov *et al*. (2021), Belle and Papantonis (2021), McDermid *et al*. (2021), Langer *et al*. (2021), Vilone and Longo (2021), Minh *et al*. (2022), and Speith (2022).

Figure 2. Conceptual taxonomy of XAI

## 2.2    *Interpretability*

Interpretability refers to the passive property of a learning model based on a *priori* knowledge. The goal of interpretability is to ensure that a given model makes sense to a human observer using language that is meaningful to users (Flock *et al*., 2021). The "success of this goal is tied to the cognition, knowledge, and biases of the user" (Gilpin *et al*., 2018: p.81). Thus, interpretability is viewed as being logical and providing a transparent outcome; that is, it is readily understandable (Marr, 1982). Adding to the mix, Riberio *et al*. (2018) suggest that an interpretable model should also be presented to the user with visual or textual artifacts to aid transparency. In Table 1, a selection of popular DL and ML models utilized in construction are examined in accordance with three dimensions of transparency identified by Lipton (2018). Each of these dimensions is discussed below.

### 2.2.1    Transparency

Underpinning the need for transparency is the need for fair and ethical decision-making. It is widely agreed that algorithms need to be efficient. They also need to be transparent and honest and be open to ex-post and ex-ante inspection (Goodman and Flaxman, 2017). Therefore, transparency aims to overcome the black-boxness associated with AI-based models (Lipton, 2018). Transparency is defined as the "ability of a human to comprehend the (ante-hoc) mechanism employed by a predictive model" (Sokol and Flach, 2022: p.5). The dimensions of *transparency* identified in Table 1 are now examined:

Table 1. Classification of DL and ML models in accordance with their level of explainability

| | Level of Explainability | | | |
|---|---|---|---|---|
| **Model** | **Simulatability** | **Decomposability** | **Algorithmic Transparency** | **Post-hoc Explainability** |
| Linear/Logistic Regression | Predictors are human-readable, and interactions among them are kept to a minimum | Variables are still readable, but the number of interactions and predictors involved in them has grown to force decomposition | Variables and interactions are too complex to be analyzed without mathematical tools | Not required |
| Decision Trees | A human can simulate and obtain the prediction of a decision tree on their own without requiring any mathematical background | The model comprises rules that do not alter data whatsoever and preserve their readability | Human-readable rules that explain the knowledge learned from data and allow for a direct understanding of the prediction process | Not required |
| K-Nearest Neighbor | The complexity of the model (number of variables, their understandability, and the similarity measure under use) matches naive human capabilities for simulation | The number of variables is too high and/or the similarity measure is too complex to be able to simulate the model entirely, but the similarity measure and the set of variables can be decomposed and analyzed separately | The similarity measure cannot be decomposed, and/or the number of variables is so high that the user has to rely on mathematical and statistical tools to analyze the mode | Not required |
| Rule-based Learners | Variables included in rules are readable, and the size of the rule set is manageable by a human user without external help | The size of the rule set becomes too large to be analyzed without decomposing it into small rule chunks | Rules have become so complicated (and the rule set size has grown so much) that mathematical tools are needed for inspecting the model behavior | Not required |
| General Additive Models | Variables and the interaction among them as per the smooth functions involved in the model must be constrained within human capabilities for understanding | Interactions become too complex to be simulated, so decomposition techniques are required to analyze the model | Due to their complexity, variables, and interactions cannot be analyzed without the application of mathematical and statistical tools | Not required |
| Bayesian Models | Statistical relationships modeled among variables and the variables themselves should be directly understandable by the target audience | Statistical relationships involve so many variables that they must be decomposed in marginals to ease their analysis | Statistical relationships cannot be interpreted even if already decomposed, and predictors are so complex that models can only be analyzed with mathematical tools | Not required |
| Random Forest | - | - | - | Usually, Model simplification or Local explanations techniques (e.g., counterfactuals[*]) |
| Support Vector Machine | - | - | - | Usually, Model simplification or Local explanations techniques |
| Multi-layer Neural Networks | - | - | - | Usually, Model simplification, Feature relevance, or Visualization techniques |
| Convolutional Neural Networks | - | - | - | Usually, Feature relevance or Visualization techniques |
| Recurrent Neural Networks | - | - | - | Usually, the Feature relevance technique and local explanations |

[*] Counterfactual explanations describe predictions of individual instances. The 'event' is the predicted outcome of an instance. The 'causes' are the feature values of this instance, which form the input to the model, causing a certain prediction (Dandl and Molnar, 2022).

Adapted from: Arrieta *et al*. (2020: p.90)

1. *Simulatability* at the model level – If a model is transparent, then a person understands the entire model, which implies it is uncomplicated (Murdoch *et al*., 2019). So, for a model to be understood, a person should be able to take the input data juxtaposed with the parameters of models and produce a prediction. As Bell and Papantonis (2021) noted, only simple and compact models fit into this mold. Imhog *et al*. (2018) and Huber and Imhof (2019), using bid distribution data from the Swiss construction sector, showed that sparse linear models produced by lasso logit regression could predict the presence of collusion. The research of Imhog *et al*. (2018) and Huber and Imhof (2019) reinforces the claim by Tibshirani (1996) that "sparse linear models, as produced by lasso regression, are more interpretable than dense linear models learned on the same inputs" (Lipton, 2018: p.13). The trade-off between model size and the computation to produce a single prediction (inference) can vary between models. But, the point at which this trade-off needs to occur is subjective, rendering linear models, rule-based systems, and decision trees to become uninterpretable.

2. *Decomposability* at the level of individual components whereby a model is broken down into parts (i.e., inputs, parameters, and computations) and then explained (Bell and Papantonis, 2021). Here inputs need to be individually interpretable, but this is not always possible, eliminating several models with highly engineered or unknown features. While sometimes intuitive, the weights of a linear model can be weak regarding feature selection and pre-processing. For example, the coefficient related to the association between the number of bidders in an auction and collusion may be positive or negative, depending on whether the feature set includes screen indicators such as pre-tender estimate, the value of the bid, and prices offered by bidders (García Rodríguez *et al*., 2022; Signor *et al*., 2023).

3. *Algorithmic* at the level of the training algorithm. This level of transparency aims to understand the procedure a model goes through to generate its output. For example, a model that classifies instances based on some similarity measure (such as K-Nearest Neighbor (K-NN), a non-parametric, supervised learning classifier) satisfies this property since its procedure is explicit (Belle and Papantonis, 2021). In this case, the datapoint similar to the one under consideration is found and assigned to the former, the same class as the latter (Belle and Papantonis, 2021). In the case of DL methods, a lack of algorithmic transparency prevails. Indeed, there exists a lack of understanding of how they work even though their heuristic optimization procedures are highly efficient. Moreover, there is no guarantee *a priori* that they will work on new problems of a similar ilk.

The three dimensions of transparency span diverse processes fundamental to predictive modeling. While these dimensions provide technical insights into the functioning of AI-based systems, their understanding offers a comprehensive overview of the issues that need to be considered to address transparency. In the AI based reviews undertaken in construction, no attention has been given to the three dimensions of transparency identified. Instead, emphasis has been placed on identifying areas where AI-based systems have been applied and where they could be used in the future without explaining what is required from models and how a model functions (e.g., Darko *et al*., 2020; Debrah *et al*., 2022; Zhang *et al*., 2022).

### 2.2.2 Completeness

The objective of completeness is to accurately assess the functioning of a model. Hence, an explanation is deemed complete when it allows the behavior of a system to be anticipated under different conditions (Gilpin *et al*., 2018). For example, in the context of ML and its use to detect collusion in public auctions, a complete explanation can be provided by presenting the mathematical operation and parameters (García-Rodríguez *et al*., 2022). The challenge that confronts XAI is providing an explanation that is both interpretable and complete (Gilpin *et al*., 2018: Arrieta *et al*., 2020; Longo *et al*., 2020; Flock *et al*., 2021).

As pointed out by Gilpin *et al*. (2018), "the most accurate explanations are not easily interpretable to people; and conversely, the most interpretable descriptions often do not provide predictive power" (p.81). The issue here is that the principle of Occam's Razor (i.e., inductive bias) often manifests when an interpretable system is evaluated with a preference given for simpler descriptions (Herman, 2017). The result is the tendency to create *persuasive* as opposed to transparent systems (Herman, 2017), which is the case for many studies of AI in construction.

For example, Flyvbjerg *et al*. (2022) developed an AI-based model based on a combination of algorithms (e.g., Random Forest, RF, and DL) to predict when construction projects were beginning to experience deviations in their cost performance and determine their final outturn cost. With such algorithms, issues of interpretability immediately come to the fore, as predicting the cost performance of a project is a wicked problem (Love *et al*., 2022b). In this case, the decomposability of the model and its algorithmic transparency require a complete explanation writ large. Moreover, *why* and *how* the DL algorithm predicts an average error of ± 8% with *only* a sample of 849 projects is not explained. Without an explanation (e.g., explainable data), the AI approach proposed by Flyvbjerg *et al*. (2022) should be treated with caution.

Simplified descriptions of complex systems are often presented in the construction literature to increase trust in the outputs produced. Still, "if the limitations of the simplified description cannot be understood by users" or the explanation is deliberately "optimized to hide undesirable attributes of the systems," then this constitutes unethical practice (Gilpin *et al*., 2018: p. 81). Such explanations are misleading and can lead to erroneous or precarious decisions. In addressing this problem, Gilpin *et al*. (2018) suggest that explanations must consider a trade-off between *interpretability* and *completeness*. In this case, Gilpin *et al*. (2018) recommend that a high-level description of the completeness of AI models should be provided at the expense of interpretability.

## 3.0    Explainability Methods

Machine learning algorithms vary in performance levels and are less accurate than DL systems (Angelov *et al*., 2021). So, the choice of algorithm to apply for a given problem is based on a trade-off between performance and explainability (Arrieta *et al*., 2020; Angelov *et al*., 2021; Sokol and Flach, 2022). A cursory review of the literature on AI in construction reveals that little attention is given to this performance-explainability trade-off. As can be seen in Figure 2, there are typically two types of XAI models:

1.    *Transparent (design)* is where a model is explained during its training. That is, the intrinsic architecture of models satisfies one of three transparency dimensions, as discussed in Section 2. As noted in Table 1, examples of transparent models, among others, are linear regression, decision trees, K-NN, rule-based learners, general additive and Bayesian learners; and

2.    *Opaque*, where the model is explained after it has been trained. Methods such as RF, Support Vector Machine (SVM), Convolutional Neural Networks (CNN), Multi-layer

Neural Networks (MNN), and Recurrent Neural Networks (RNN) identified in Table 1 are used instead of transparent methods due to unsatisfactory performance. Yet, opaque models are the epitome of a black box and thus require post-hoc explainability to try and generate explanations, though they have already been trained (Yamashita *et al*., 2018; Hasenstab *et al*., 2023). As explanators or explainers are developed, greater insights into the inner structure, mechanisms, and properties of opaque models will be acquired.

Notably, there are distinct differences between *model-agnostic* and *model-specific* methods. A model-agnostic method works for all models. In contrast, the model-specific methods only work for specific models such as the DL algorithms, SVM, and RF (Speith, 2022). The next sections of this paper briefly examine the nature of the transparent and opaque methods used to support XAI.

### 3.1   *Transparent Models*

As seen in Figure 2, several transparent models frequently used in the literature and widely applied in construction are identified. It is beyond the scope of this paper to provide a detailed review of studies that examine transparent models in construction. However, Xu *et al*. (2021) provide a comprehensive review of shallow[1] and deep[2] learning models and their application to construction. Each of the models identified in Figure 2 and Table 1 is briefly examined, with reference being made to examples from construction (Arrieta *et al*., 2020; Belle and Papantonis, 2021):

---

[1] A shallow model refers to learning from data described by predefined features. In the case of a shall neural network only one hidden layer of nodes exists.
[2] Deep learning is a form of a neural network with three or more layers. The neural networks aim to simulate the behavior of the human brain – albeit from matching its ability – allowing it to learn from large amounts of data

- *Linear/Logistic Regression*: Such models are used to predict a continuous or categorical dependent variable that is dichotomous. The assumption underpinning such models is that there is a linear dependence between the predictors (independent) and predicted variables. Consequently, there is no flexibility to fit data – they are stiff models. The explainability of these models is directly related to who is interpreting the output. So, in the case of laypersons, variables need to be understandable. While these models adhere to the three dimensions of transparency denoted in Table 1, there may be a need for post-hoc explainability using visual techniques, which is discussed in Section 3.3. Examples of the use of linear and logistic regression abound in the construction literature and, while simple, have demonstrated high levels of accuracy (Rezaie *et al*., 2022).

- *Decision Trees*: These models support classification and regression problems and are often called *Classification and Regression Trees* (CART). In their simplest form, decision trees are simulatable models. They can also be decomposable or algorithmically transparent due to their properties. A simultable decision tree is typically small. Its features and meaning are generally understandable and thus manageable by a human user. As trees become larger, they transform into decomposable ones as their size hinders their complete evaluation by an end-user. Any further increases in the size of a tree can result in complex feature relations, rendering it algorithmic transparent. A cursory review of the construction literature reveals that decision trees, in various guises, have been applied to an array of applications. For example, Shin *et al*. (2012) apply decision trees to help select formwork methods. Due to their off-the-shelf transparency, decision trees are often used to create a decision support system (DSS) (Arrieta *et al*., 2020). For example, You *et al*. (2018) use decision trees to develop a DSS to classify design options for roadways of coal mines. From a mathematical perspective, a decision tree has low bias and high variance. Averaging the results of several decision trees reduces the variance while maintaining a low bias. The combining of trees is known as the Ensemble Tree method (an opaque model). Here, a group of weak learners comes together to form a strong learner, which improves its predictive performance. Ensemble techniques typically used in construction are *Bagging*[3] (also known as Bootstrap Aggregation with the RF model being an extension) and *Boosting*[4] (an extension is Gradient Boosting). Examples where the ensemble method has been applied in construction vary and include:

o   Predicting cost overruns (Williams and Gong, 2014);

o   Matching expert decisions in concrete dispatching centers (Maghrebi *et al*., 2016); and

o   Developing a cost-effective wireless measurement method using ultra-wideband radio technology is enhanced by an extreme gradient boosting tree (Liu *et al*., 2021).

However, transparency is lost when decision trees are combined and integrated with other ML techniques. In such cases, post-hoc explainability is required (Section 3.3).

- *K-Nearest Neighbor*: This ML method deals with classification problems simply and methodically. It is a non-parametric, supervised learning classifier that uses proximity to make classifications or predictions about the grouping of an individual data point. Thus, it does not make any assumptions about the underlying data and does not learn from a training set but instead stores the dataset. At the time of classification, it acts on the dataset. As the K-NN models rely on the notion of distance and similarity between examples, they can be tailored to a specific problem enabling model explainability. In addition to being simple to explain, K-NNs are interpretable. The reasons for classifying a new sample within a group and how these predictions materialize when the number of neighbors K ± can be examined by users. The K-NN has been widely applied to classification problems and combined with other ML techniques in construction, including:

o    The development of a K-NN knowledge-sharing model that can be used to share links of the behaviors of similar project participants that have experienced disputes as a result of change orders (Chen *et al*., 2008)

o    Case retrieval in a case-based reasoning cycle to search for similar plans to create a new technical plan for constructing deep foundations (Zhang *et al*., 2017); and

o    The selection of the most informative features is part of a process to automate the location of the damage on the trusses of a bridge (Parisi *et al*., 2022).

---

[3] Bagging is used to reduce the variance of a decision tree. Several subsets of data from a training sample are randomly selected and replaced.
[4] Boosting creates a collection of predictors. Learners are learned sequentially with early learners fitting simple models to data and then performing analysis to identify errors.

Noteworthy, the use of complex features and/or distance functions can hinder the model decomposability of a K-NN, "which can restrict its interpretability solely to the transparency of its algorithmic operations" (Arrieta *et al*., 2020: p.91).

- *Rule-based Learners*: This class of models refers to those that generate rules to characterize data they intend to learn from. Rules can take the form of conditional *if-then* rules or more complex combinations. Fuzzy-rule-based systems (FRBSs) are rule-based learners. They are models based on fuzzy sets that express knowledge in a set of fuzzy rules to address complex real-world problems where uncertainty, imprecision, and non-linearity exist. As FRBSs operate in a linguistic domain, they tend to be understandable and transparent as rules are used to explain predictions. A detailed review of studies that have applied FRBSs in construction can be found in Robinson-Fayek (2020). Besides FRBSs, rule-based methods have been extensively used for knowledge representation in expert systems in construction for several decades (Adeli, 1988; Minkarah and Ahmad, 1989; Levitt and Kartam, 1990; Irani and Kamal, 2014). Though, a major setback with rule generation methods is the number and length of the rules generated. At the same time, as the number of rules increases, the performance of systems will improve but at the expense of interpretability. Moreover, specific rules may have many antecedents or consequences, making it difficult for a user to interpret its outputs. Hence, the greater the coverage or specificity of a system, the closer it is to be algorithmically transparent. In sum, rule-based learners are generally able to be interpreted by users.

- *General Additive Model (GAM)*: These are linear models where the outcome is a linear combination of functions based on the input features. Thus, the value of the variable to be predicted is given by the aggregation of several unknown smooth functions defined for the predictor variables. The GAM aims to infer the smooth functions whose aggregate composition approximates the predicted variable. This structure is readily interpretable, as it enables a user to verify the importance of each variable and how it affects the expected output. Within the construction literature, the use of GAMs has been limited to applications such as assessing the productivity performance of scaffold construction (Liu *et al*., 2018). Thus, they have been generally used to determine risk in other fields, such as finance, health, and energy (Arrieta *et al*., 2020). Visualization methods, such as *dependency plots*, are typically used to enhance the interpretability of models (Friedman and Meulman, 2003). Similarly, GAMs are deemed to be simultable and decomposable if the properties mentioned in their "definitions are fulfilled", though this depends on the extent of modifications made to the baseline model (Arrieta *et al*., 2020: p.91)

- *Bayesian Models*: These models tend to form a probabilistic directed (acyclic) graphical model linked to represent conditional probabilities between a set of variables. As a clear and distinct connection between variables and probabilistic relationships is displayed, they have been used extensively used in construction in areas such as:

  o The development of site-specific statistics that model bias to enable the design of cost-effective pile foundations (Zhang *et al*., 2020);

  o Modeling the risk of collapses during the construction of subways (Zhou *et al*., 2022); and

  o Surface settlement prediction during urban tunneling (Kim *et al*., 2022).

  Such models are transparent as they are simulatable, decomposable, and algorithmically transparent. However, when models become complex, they tend to become only algorithmically transparent.

The transparent models briefly described above are prevalent in construction. Still, issues associated with the precepts of XAI have seldom been given any serious consideration while justifying their relevance to practice.

## 3.2 *Opaque Models*

Opaque models generally outperform transparent ones in terms of their predictive accuracy (Angelov *et al*., 2021). Hence, their widespread appeal and popularity in construction.. Common opaque models used in construction are now examined:

- *Random Forest*: This ML model improves the accuracy of single decision trees, which often suffer from overfitting and, thus, poor generalizations. Random Forest addresses the issue of overfitting by combining multiple trees to reduce the variance in models produced and thus improve its generalization. As mentioned in Section 3.1, an RF is an extension of an Ensemble Tree and is a shallow ML model. Examples, where this algorithm has been applied in construction, include:

  - Activity recognition of construction equipment (Langroodi *et al*., 2020);
  - Predicting the cost of labor on a building information modeling project (Huang and Hsieh, 2020); and
  - The classification of surface settlement levels induced by tunnel boring machines in built-up areas (Kim *et al*., 2022).

  The so-called forest of trees produced is difficult to interpret and explain compared to single trees. As a result, post-hoc explainability is required using *local explanations* (see below).

- *Support Vector Machine*: This shallow model is a deep-learning algorithm that performs supervised learning for the classification or regression of data groups. It has a historical presence in the construction literature, as shown in the in-depth and insightful review of Garcia *et al*. (2022). Indeed, SVM models are inherently more complex than Tree Ensembles and RF, and considerable effort has gone into mathematically describing their internal functioning. However, these models require post-hoc explanations due to their high dimensionality and likely data transformations rendering them opaque and requiring explanation by simplification, local explanations, visualizations, and explanations by example (Section 3.2.1). An SVM model "constructs a hyper-plane or set of hyper-planes in a high or infinite dimensional space, which can be used for classification, regression, or other tasks such as outlier detection" (Arrieta *et al.*, 2020: p.95). Separation is considered good when the *functional margin* is a significant distance from the nearest training data point for a given class; thus, a larger margin will lower the generalization error of the classifier. For the benefit of readers, the functional margin simply represents the lowest relative confidence of all classified points.

- *Multi-layer Neural Network*: This method, also called Multilayer Perceptron (MLP), is an integral part of DL. The DL has three hidden layers and is a typical example of a feed-forward artificial neural network (ANN). The number of layers and neurons are referred to as the hyperparameters of models, which require tuning. This tuning needs to be undertaken through a process referred to as cross-validation. A detailed review of ANNs in construction can be found in Xu *et al*. (2022). Specific examples of studies using MNN in construction include:

  o A comparison of cost estimations of public sector projects using MLP and a radial basis function (Bayram *et al*., 2016);
  o Optimizing material management in construction (Golkhoo and Moselhi, 2019); and
  o The design exploration of quantitative performance and geometry for arenas with visualizations generated to interpret the results (Pan *et al.*, 2020).

Despite the widespread use of MNN algorithms in construction, they are prone to the vanishing gradient problem. This situation arises when the deep multilayer feed-forward network cannot propagate useful gradient information from the output end of the model back to the layers near its input. The questionable explainability (i.e., black box) of these models also hinders their application in practice. Accordingly, multiple explainability techniques are needed to acquire an understanding of their operation and outcomes, which include model simplification methods (e.g., rule extraction and DeepRED algorithm) (Sato and Tsukimoto, 2001; Zilke *et al*., 2016), text explanations, local explanations, and visual explanations

- *Convolutional Neural Network*: These are state-of-the-art models for computer vision and are used for various purposes, such as image classification, object detection, and instance segmentation. A CNN is designed to automatically and adaptively learn spatial hierarchies of features through backpropagation using multiple building blocks, such as convolution, pooling, and fully connected layers. The structure of a CNN is comprised of complex internal relations and, thus, is difficult to explain. Nonetheless, research is making progress toward explaining what a CNN can learn with increased emphasis placed on (Arrieta *et al*., 2020): (1) understanding their decision process by mapping back the output in the input space to determine the parts of the input that are discriminative on the output (Zeiler *et al*., 2011; Bach *et al*., 2015; Zhou *et al*., 2016); and (2) delving inside the network to interpret how the intermediate layers view the external environment (e.g., determining what neurons have learned to detect) (Mahendran and Vedaldi, 2015; Nguyen *et al*., 2016). Despite the need for post-hoc explainability, there has been a significant upsurge in using CNNs and computer vision in construction, with numerous review papers published (Martinez *et al*., 2019; Zhong *et al*., 2019; Paneru and Jeelani, 2021). Additionally, several domain-specific reviews have come to the fore, which include safety assurance (Fang *et al*., 2020), interior progress monitoring (Ekanayake *et al*., 2021), and structural crack detection (Ali *et al*., 2022).

- *Recurrent Neural Network*: While CNNs dominate the visual domain, the RNN is a feed-forward model for predictive problems with predominately sequential data that uses patterns to forecast the next likely scenario (e.g., time series analysis). The data used for RNNs tends to exhibit long-term dependencies that ML models find difficult to capture due to their complexity, which impacts training (e.g., vanishing and exploding gradients) and their computation (Pascanu *et al*., 2013). However, RNNs can retrieve time-dependent relationships by "formulating the retention of knowledge in the neuron as another parametric characteristic that can be learned from the data" (Arrieta *et al*., 2020: p.98). The explainability of RNNs remains a challenge, despite significant attention focusing on understanding what a model has learned and the decisions that they make (Arras *et al*., 2017; Belle and Papantonis, 2021). Regardless of the difficulties associated with the explainability of RNNs, they are popular models in construction. For example, they have been used to:

o        Develop a control method for a constrained, underactuated crane (Duong *et al*., 2012)

o        Create a real-time prediction of tunnel-boring machine operating parameters (Gao *et al*., 2019); and

o        Predict the control of slurry pressure in shield tunneling (Li and Gong, 2019).

With explainability, interpretability, and transparency needing to be addressed with RNNs, caveats on their outputs must be explicit. By being extra vigilant with their use and, in addition to post-hoc explainability, methods such as ablation, permutation, random noise, and integrated gradients within given contexts can be applied to ensure compliance with the goal of XAI.

### 3.2.1   Post Hoc Explainability

Black-box (i.e., opaque) models are not interpretable by design, but they can be interpreted after training without sacrificing their predictive performance (Lipton, 2018; Speith, 2022). Thus, post-hoc explainability "targets models that are not readily interpretable by design" by augmenting their interpretability, as denoted in Table 1 (Lipton, 2018: Arrieta *et al*., 2020: p.88). Even though post-hoc explanations are unable to describe exactly how ML models function, they can provide some useful information that can help end-users understand their outputs using the following techniques (Lipton, 2018; Arrieta *et al*., 2020; Belle and Papantonis, 2021; Hussain *et al*., 2021; Speith, 2022):

- *Text explanations* produce explainable representations utilizing symbols, such as natural language text. Symbols can be used to represent the rationale of the algorithm by semantically mapping from model to symbol. The work of Zhong *et al*. (2021) is relevant here, as they use text to explain the decisions of a latent factor model. Zhong *et al*. (2020) simultaneously train a Latent Dirichlet Allocation (LDA) model and a CNN to garner insights into the causal nature of accidents from text narratives alongside visual information.

- *Visual explanations* present qualitative outputs denoting the behavior of models. A particular case in point is the research of Miller (2019), which used visualizations as part of their study of explainable ML to understand the behavior of energy performance in non-residential buildings. A popular method is to visualize high-dimensional presentations with t-distributed Stochastic Neighbor Embedding (t-SNE). The t-SNE is a technique that visualizes high-dimensional data by giving each point a location in a two or three-dimensional map. For example, while examining concrete cracks, Geeta and Sim (2022) use the metadata within a one-dimensional DL model to visualize the knowledge transfer between its hidden layers using t-SNE. Visualizations may also be used with other techniques to enhance our understanding of complex interactions with the variables included in a model (Zhong *et al*., 2020).

- *Local explanations* explain how a model operates in a certain area of interest. While examining how a neural network learns is challenging, the behavioir of the model can be explained locally. For example, an approach used for CNNs is saliency mapping (Simonyan *et al*., 2013). In this instance, salience maps aim to define where a particular region of interest (RoI) (i.e., a place on an image that is searched for something) is different from neighbors with respect to image features. Typically, the gradient of the output corresponding to the correct class for a given input vector is taken (Lipton, 2018). So, for images, this "gradient is applied as a mask highlighting regions of the input that, if changed, would most influence the output" (Lipton, 2018: p.18). For example, in Fang *et al*. (2019), the RoI – at the pixel level – was the segmentation of people from structural supports across deep foundation pits from images. Thus, to determine the distance and relationship between people and the structural supports, Fang *et al*. (2019) utilized a Mask-Region-CNN. Noteworthy, a salience map only provides a local explanation, meaning a different saliency map can emerge if one pixel is changed.

- *Explanations by example,* extract representative instances from a training dataset to show how a model functions. The training data needs to be in a format understandable by people, such as images, as arbitrary vectors can contain hundreds of variables, making it difficult to unearth information. Training a CNN or latent variable model (e.g., LDA) for a discriminative task[5] provides access to both predictions and learned representations. In this instance, besides generating a prediction, the "activations of the hidden layers" to identify K-NN based on the proximity in the space-learned model can be used (Lipton,

---

[5] A discrimination task is a discrimination test in which the two test samples, A and B (one of which is the control sample, the other a modified sample), are presented to the participant first followed by sample X (Yotsumoto *et al*., 2009)

2018: p.29). In the construction literature, the use of CNNs has been widespread to examine learned representations of words after training with a word2vec model (Baek *et al*., 2021). For example, Pan *et al*. (2022) developed a consortium blockchain network to ensure information authentication and security for equipment using a word22vec and CNN to automatically identify keywords and categories from inspection reports. In accordance with the procedure presented in Mikolov *et al*. (2013), the model developed by Pan *et al*. (2022) was trained for discriminative skip-gram[6] prediction to examine the relationships it has learned, while the nearest neighbors of words are based on distances calculated in the latent space.

- *Explanations by simplification* denote those techniques (e.g., rule-based learner, decision trees, and knowledge distillation[7]) used to rebuild a system based on an explained trained model rendering it easier to interpret. The newly developed model aims to optimize its resemblance to its antecedence functioning to obtain a similar performance level. An issue here is that the simple model needs to be flexible so that it can approximate the model accurately (Belle and Papantonis, 2021); and

- *Feature relevance explanation* techniques (e.g., Local Interpretable Model-agnostic Explanations[8] (LIME) and Shapley Additive Explanation[9] (SHAP)) aim to explain the quality of the inner functioning of an MLs learning model and decision by calculating the influence of each input variable and produce relevant scores. These scores quantify the *sensitivity* of a feature on the output of a model. The scores alone do not provide a complete explanation but rather enable insights into the functioning and reasoning to be garnered from the model. Most of the commonly used feature importance techniques are Mean Decrease Impurity (MDI), Mean Decrease Accuracy (MDA), and Single Feature Importance (SFI). The techniques aim to generate a feature score directly proportional to its effect on the overall predictive quality of the ML model.

---

[6] Skip-gram is an unsupervised learning technique used to find the most related words for a given word. Skip-gram is used to predict the context word for a given target word.
[7] Knowledge distillation is one form of the model-specific XAI. Here this relates to eliciting knowledge from a complicated to a simplified model.
[8] The LIME is a model explanation algorithm that provides insights into how much each feature contributed to the outcome of a ML model for an individual prediction
[9] The SHAP is a model explainability algorithm that operates on the single prediction level rather than the global ML model level.

As an area of research, XAI is growing exponentially, with the number of papers exceeding hundreds annually (Arrieta *et al*., 2020; Zhou *et al*., 2022; Yang *et al*., 2022). Taking stock of new developments in XAI is a challenge. However, the precepts, methods, and post-hoc explanations discussed above provide a basis for understanding the specific facets and requirements of XAI. Against the contextual backdrop provided, the opportunities for XAI research in construction are next examined.

## 4.0   Opportunities for Research in Construction

Unquestionably XAI research is accelerating at an unprecedented rate of knots, so keeping abreast of the latest developments is challenging. There are several key areas where research opportunities in construction can be harnessed and the benefits of XAI realized. It is suggested that immediate research opportunities reside around (1) stakeholder desiderata; and (2) data and information fusion. All in all, XAI provides a fundamental step toward establishing fairness and addressing bias in algorithmic decision-making (Mougan *et al*., 2021).

The language surrounding explainability is inconsistent; a recurring theme that echoes throughout the extant computer science and information systems literature (Murdoch *et al*., 2019; Arrieta *et al*., 2020; Belle and Papantonis, 2021; Mougan *et al*., 2021; Hussain *et al*., 2021; Speith, 2022; Zhou *et al*., 2022; Yang *et al*., 2022). Explicit explainability goals, desiderata, and methods for evaluating the quality of explanations are evolving but are also challenged when they come to fruition. There is a consensus that there is a lack of rigor in evaluating explanation methods (Doshi-Velez and Kim, 2017; Mougan *et al*., 2021). A significant factor contributing to this ongoing predicament is that evaluation criteria tend to be based solely on AI engineers' views rather than *stakeholder desiderata,* where their specific interests, goals, expectations, and demands for a system within a given context are considered (Amarasinghe *et al*., 2021; Langer *et al*., 2021; Love *et al*., 2022c).

At face value, the absence of agreed evaluation criteria is no doubt frustrating for stakeholders of systems. However, it is suggested that this robust and healthy challenge of ideas will help the field of XAI move forward quickly rather than stagnate progress. Advancements in the technical aspects of XAI (e.g., development of algorithms and evaluation metrics) should not be an immediate area of concern for researchers in construction, as those in the field of computer science and engineering is attending to these issues.

Similarly, the challenges associated with developing legal and regulatory requirements are beyond the remit of researchers in construction, though being cognizant of developments and how they impact practice is an area that warrants investigation. For example, the European Union has proposed a regulatory framework for AI to provide developers, deployers, and users with precise requirements and obligations regarding its specific use (European Commission, 2020).

Many DL and ML algorithms exist, which can form part of a researcher's XAI toolbox. Markedly, many construction researchers have acquired the skills and knowledge to use DL and ML models effectively. The eagerness of researchers to apply new algorithms to a problem is evident in the literature. No sooner than a new DL or ML model is developed, and a flurry of academic articles is published in construction journals to show it works for an artificially developed problem. However, rarely does research in construction demonstrate the *utility* of DL or ML models in practice and evaluate whether systems do what they are supposed to do (Love *et al*., 2022c). So going forward, and with the emergence of XAI, it is strongly recommended that research in construction begin to focus on developing AI-enabled solutions based on stakeholder desiderata. After all, addressing stakeholder desiderata is the main reason for the rising popularity of XAI (Langer *et al*., 2021).

### 4.1    Stakeholder Desiderata

Stakeholders are "the various people who operate AI systems, try to improve them, are affected by decisions based on their output, deploy systems for everyday tasks, and set the regulatory frame for their use" (Langer *et al*., 2021: p.3). Within the context of construction, stakeholders will include the client, project teams, subcontractors, regulatory bodies, and the public. Thus, Langer *et al*. (2021) maintain that the success of XAI depends on the satisfaction of stakeholders' desiderata, comprising of the *epistemic* and *substantial* facets.

Ensuring the desideratum of satisfaction, identified in Table 2, is managed and maintained will help ensure stakeholders are satisfied with adopting and implementing AI. To clarify the nature of stakeholders identified in Table 2 and provide a context to construction, an example of a deployer could be the construction organization, a regulator, the client, the developer, an independent software provider, and a user who is a representative from a site management team of a project.

Each stakeholder has a role to play in ensuring the success of an AI system. Ensuring their divergent goals, needs, and interests do not compromise the integrity of the AI system will require constant attention and management, especially as new desiderata emerge. Determining who and how the stakeholder desiderata are managed and evaluated is an issue that will need to be examined in construction. Furthermore, the solicitation of explainability needs from stakeholders will also need to be considered. For example, a construction manager (i.e., the end-user) using a safety risk-assessment AI would most likely want an overview of the system during its onboarding stage. Also, they may like to delve deeper into the reasoning for a particular series of work-related risks during the planning process.

Examples of questions, modified from Liao and Varshney (2022), which end-users can ask (e.g., construction manager), in addition to those in Figure 1, can be seen in Figure 3. Questions can also be used to guide the choices of XAI techniques. For example, a feature-importance explanation can answer *why* questions, while a counterfactual can answer *how* questions. A questioning mindset will naturally invite conversation, understanding, and knowledge growth which are essential for explainability.

Stakeholders in construction will naturally require AI systems to have specific properties that enable them to be transparent and useable. Though, they must be educated about the *what*, *why*, *how,* and *when* of AI. Similarly, they will "want to know or be able to assess whether a system (substantially) satisfies a certain desideratum", such as those identified in Table 2 (Langer *et al*., 2021). Thus, the epistemic facet of trustworthiness is satisfied for a stakeholder, such as a construction manager, if they can assess or know if the output of a system is accurate and reliable. Developing successful explanation processes is an area of research the construction community will need to address for XAI in the foreseeable future. This point provides a segue for discussing the next proposed opportunity for XAI research.

Table 2. Examples of desideratum that contribute to their satisfaction in XAI systems

| Desideratum | Brief Description | Stakeholder |
|---|---|---|
| Acceptance | Improve acceptance of systems | Deployer, Regulator |
| Accountability | Provide appropriate means to determine who is accountable | Regulator |
| Accuracy | Assess and increase the predictive accuracy of the system | Developer |
| Autonomy | Enable humans to retain their autonomy when interacting with a system | User |
| Confidence | Make humans confident when using a system | User |
| Controllability | Retain (complete) human control concerning a system | User |
| Debuggability | Identify and fix errors and bugs | Developer |
| Education | Learn how to use a system and its peculiarities | User |
| Effectiveness | Assess and increase the effectiveness of a system; work effectively with a system | Developer, User |
| Efficiency | Assess and increase the efficiency of a system; work efficiently with a system | Developer, User |
| Fairness | Assess and improve the fairness of a system | Affected, Regulator |
| Informed consent | Enable humans to give their informed consent concerning the decisions of a system | Affected, Regulator |
| Legal compliance | Assess and increase the legal compliance of a system | Deployer |
| Morality/Ethics | Assess and increase compliance with the moral and ethical standards of a system | Affected, Regulator |
| Performance | Assess and improve the performance of a system | Developer |
| Privacy | Assess and improve the privacy practices of a system | User |
| Responsibility | Provide appropriate means to let humans remain responsible or increase perceived responsibility | Regulator |
| Robustness | Assess and increase the robustness of a system (e.g., against adversarial manipulation) | Developer |
| Safety | Assess and increase the safety of a system | Developer, User |
| Satisfaction | Have satisfying systems | User |
| Security | Assess and increase a systems security | All |
| Transferability | Make the learned model of a system transferable to other contexts | Developer |
| Transparency | Have transparent systems | Regulator |
| Trust | Calibrate appropriate trust in the system | User, Developers |
| Trustworthiness | Assess and increase the trustworthiness of a system | Regulator |
| Usability | Have usable systems | User |
| Usefulness (Utility) | Have useful systems | User |
| Verification | Be able to evaluate whether the system does what it is supposed to do | Developer |

Adapted from Langer *et al.* (2021: p.5)

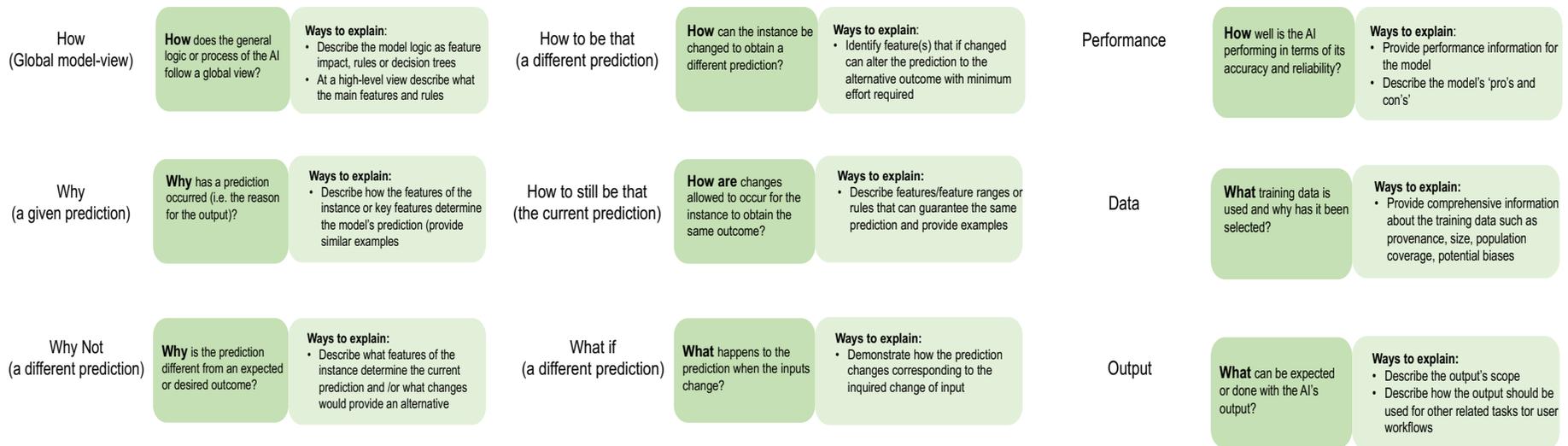

How
(Global model-view)

**How** does the general logic or process of the AI follow a global view?

**Ways to explain:**
• Describe the model logic as feature impact, rules or decision trees
• At a high-level view describe what the main features and rules

How to be that
(a different prediction)

**How** can the instance be changed to obtain a different prediction?

**Ways to explain:**
• Identify feature(s) that if changed can alter the prediction to the alternative outcome with minimum effort required

Performance

**How** well is the AI performing in terms of its accuracy and reliability?

**Ways to explain:**
• Provide performance information for the model
• Describe the model's 'pro's and con's'

Why
(a given prediction)

**Why** has a prediction occurred (i.e. the reason for the output)?

**Ways to explain:**
• Describe how the features of the instance or key features determine the model's prediction (provide similar examples

How to still be that
(the current prediction)

**How are** changes allowed to occur for the instance to obtain the same outcome?

**Ways to explain:**
• Describe features/feature ranges or rules that can guarantee the same prediction and provide examples

Data

**What** training data is used and why has it been selected?

**Ways to explain:**
• Provide comprehensive information about the training data such as provenance, size, population coverage, potential biases

Why Not
(a different prediction)

**Why** is the prediction different from an expected or desired outcome?

**Ways to explain:**
• Describe what features of the instance determine the current prediction and /or what changes would provide an alternative

What if
(a different prediction)

**What** happens to the prediction when the inputs change?

**Ways to explain:**
• Demonstrate how the prediction changes corresponding to the inquired change of input

Output

**What** can be expected or done with the AI's output?

**Ways to explain:**
• Describe the output's scope
• Describe how the output should be used for other related tasks for user workflows

Figure 3. Examples of XAI questions for end-users

Evaluation frameworks for XAI are yet to be developed, albeit with an empirical foundation (Love *et al*., 2022c). Thus, if construction organizations are to benefit from XAI, then evaluation frameworks that include measurable outcomes for domain-specific applications for stakeholder desiderata will need to be cultivated.

Examples of domains receiving attention from AI engineers and software developers include generative design, management of safety and quality, off-site construction, real-time monitoring, and plant and equipment security. As mentioned already, stakeholders rarely have, if at all, been involved in developing these applications. Without the input of stakeholders, the benefits will not be realized (Love and Matthews, 2019).

Determining where AI will provide a return on investment requires construction organizations to comprehend and interpret how systems work and have confidence in the outputs (or recommendations). Despite the increasing maturity of AI as a technology, researchers and practitioners alike are still far from making many of its solutions understandable to a satisfactory level of know-why in construction. For example, how can organizations ensure an AI system performs as intended and does not negatively impact their bottom line and end-users?

Established software vendors and new entrants to the marketplace in construction will no doubt aim to tackle this issue. It will take a mixture of domain experts and developers to interpret and translate the insights from an XAI system to non-technical, understandable explanations for end-users. Moreover, while the number of AI solutions has increased in construction to reduce the need for domain experts and developers, they may not always provide explainability at the required level.

## 4.2    Data and Information Fusion

Data and information fusion is a basic capability for many state-of-the-art technologies, such as the Internet of Things (IoT), computer vision, and remote sensing. But fusion is a rather vague concept and can take several forms (Murray, 2021). For example, in the case of computer vision, the combining of features is fusion (Fang *et al*., 2020; Fang *et al*., 2021; Fang *et al*., 2022), and in remote sensing, registration is fusion (Murray, 2021).

For the purpose of this paper, data fusion refers to assembling and combining different kinds of information into a procedure yielding a single model (Chatzichristos *et al*., 2022). The process of correlating and fusing information from multiple sources generally allows more accurate inferences than those that the analysis of a single dataset can yield (Ding *et al*., 2019; Fang *et al*., 2021; Fang *et al*., 2022). Thus, data fusion can potentially enrich the explainability of ML and DL models and "compromise the privacy of the data" that has been learned (Arrieta *et al*., 2021: p.105). To this end, it is suggested that AI studies in construction, particularly those using computer vision, place emphasis on data fusion to improve interpretability and transparency explanations to help ensure the XAI precepts can be fulfilled.

Data fusion can occur at the data (i.e., raw data), model (i.e., aggregation of models), and knowledge (i.e., in the form of rules, ontologies, and the like) levels (Arrieta *et al*., 2020). As AI systems become more complex and the requirement for privacy (e.g., personal data) assured during the life-cycle of a system, big data fusion, federated learning, and multi-view learning methods for data fusion are required. It is outside the remit of this paper to discuss the various ways data fusion occurs. However, a detailed examination of why and how fusion occurs to address issues such as the IoT, privacy, and cyber-security can be found in Ramírez-Gallego *et al*. (2018), Lau *et al*. (2019), and Smirnov and Levashova (2019).

Noteworthy that data fusion at the data level has no connection to the ML models. Thus, they have no explainability. Nevertheless, the distinction between information fusion and predictive modeling with DL models is somewhat cloudy, even though their performance is improved. Again, the performance-explainability trade-off emerges. In a DL model, the first layers are responsible for learning high-level features from raw fused data. As a result, this process tightly couples the fusion with the tasks to be solved, and features become correlated (Arrieta *et al*., 2020).

Several XAI techniques (e.g., LIME and SHAP) have been proposed to deal with the correlation of features, which can explain how data sources are fused in a DL model and improve its *usability*. In the context of safety and privacy on construction sites, for example, when using computer vision, there remains little understanding as to whether the input features of a model can be inferred if a previous feature was known to be used in that model. Herein lies the final opportunity for research. Thus, it is suggested that a multi-view perspective is adopted to obtain a clearer perspective of what is going on in a model to enable XAI. In this instance, different single data sources are fused to examine how information that is not shared can be discovered.

Empirical work within construction addressing data privacy issues on-site has not been forthcoming. Yet, it is a critical barrier to applying XAI and needs to be addressed. Federated learning and differential privacy have been identified as solutions to deal with data privacy and protection and promote the acceptance of AI (Rodríguez-Barroso *et al*., 2020; Adnan *et al*., 2022). So, if headway is to be made to improve project performance (e.g., productivity, quality, and safety) using AI, then it must be explainable.

## 5.0    Conclusion

The outputs generated from the AI-based models of DL and ML have tended to be unexplainable, which can impact the decision-making efficacy of end-users. The utilization of XAI can explain why and how the output of DL and ML models are generated. Accordingly, an understanding of the functioning, behavior, and outputs of models can be acquired, reducing bias and error, and improving confidence in decision-making. While researchers and construction organizations explore how DL and ML can enhance the productivity and performance of projects and assets across their life, limited attention is being given to XAI.

Indeed, AI is shaping the future of construction and is already the main driver for emerging technologies, such as the IoT and robotics, and will continue to be a technological innovator. But, as Eliezer Yudkowsky, a computer scientist at the Machine Learning Research Institute, insightfully remarked, "by far, the greatest danger of Artificial Intelligence is that people conclude too early that they understand it." Reaching such a conclusion without explaining why and how the black box of ML and DL models function can result in precarious decision-making and outcomes. Thus, as AI becomes more intertwined with human-centric applications and algorithmic decisions become consequential, attention in construction will need to shift away from focusing solely on the predictive accuracy of DL and ML models toward their explainability.

Hence the motivation for this review has been to raise awareness about XAI and to stimulate a new agenda for AI in construction. The review developed a taxonomy of the XAI literature, comprising its precepts (e.g., explainability, interpretability, and transparency) and methods (e.g., transparent and opaque models with post-hoc explainability). Indeed, the XAI literature is complex and vast and growing exponentially, exacerbating the need to understand better why and how DL and ML models produce specific predictions. Thus, it is suggested that the proposed taxonomy can be used as a frame reference to navigate the fast and changing world of XAI.

Emerging from the review presented, opportunities for future XAI research focusing on stakeholder desiderata and data/information fusion were identified and discussed. While accountability, fairness, and discrimination were identified as key issues that require consideration in future XAI studies, the discourse associated with them is still evolving, and legal ramifications are yet to be resolved. To this end, the authors hope the XAI opportunities suggested will stimulate new lines of inquiry to help alleviate the skepticism and hesitancy toward developing an implementation pathway for AI adoption and integration in construction.

**Acknowledgments**

We would like to acknowledge the financial support of the *Australian Research Council* (DP210101281) and (DE230100123). Also, we would like to acknowledge the financial support of the Alexander von Humboldt-Stiftung, and the National Natural Science Foundation of China (Grant No. U21A20151).